\documentclass[11pt,a4paper]{article}
\usepackage[hyperref]{acl2021}
\usepackage{times}
\usepackage{latexsym}

\usepackage{microtype}

\aclfinalcopy 
\def\aclpaperid{2416} 

\usepackage[T1]{fontenc}

\usepackage[utf8]{inputenc}

\usepackage[compact]{titlesec}

\usepackage{graphicx}
\usepackage{xcolor}
\usepackage{bbold}

\usepackage{booktabs}
\usepackage{tabularx}

\definecolor{darkgreen}{rgb}{0.0, 0.4, 0.33}
\definecolor{darkorange}{rgb}{0.9, 0.45, 0.0}

\definecolor{hi-red}{HTML}{e41a1c}
\definecolor{hi-green}{HTML}{4daf4a}

\newcommand{\anthony}[1]{}
\newcommand{\avi}[1]{}
\newcommand{\heng}[1]{}
\newcommand{\hans}[1]{}

\newcommand\blfootnote[1]{%
  \begingroup
  \renewcommand\thefootnote{}\footnote{#1}%
  \addtocounter{footnote}{-1}%
  \endgroup
}

\usepackage{amsmath,amsfonts,bm}

\def\eqref#1{equation~\ref{#1}}

\def\1{\bm{1}}

\DeclareMathAlphabet{\mathsfit}{\encodingdefault}{\sfdefault}{m}{sl}
\SetMathAlphabet{\mathsfit}{bold}{\encodingdefault}{\sfdefault}{bx}{n}

\title{\systemname{}: VAriable Unified Long Text Representation for Machine Reading Comprehension}

\author{
Haoyang Wen$^{2\dag}$\thanks{\quad Work done during an internship at IBM Research AI.} ,
Anthony Ferritto$^{1\dag}$, 
Heng Ji$^{2}$\\
\textbf{Radu Florian$^{1}$,
Avirup Sil$^{1}$}\\
$^{1}$ IBM Research AI, 
$^{2}$ University of Illinois Urbana-Champaign\\
\texttt{wen17@illinois.edu, aferritto@ibm.com}\\
\texttt{hengji@illinois.edu, \{raduf, avi\}@us.ibm.com}
    }

\date{}

\newcommand{\systemname}{\textsc{Vault}}
\newcommand{\dm}{\textsc{RoBERTa$_{DM}$}}

\begin{document}
\maketitle
\begin{abstract}
Existing models on Machine Reading Comprehension (MRC) require complex model architecture for effectively modeling long texts with paragraph representation and classification, thereby making inference computationally inefficient for production use. In this work, we propose \systemname{}: a light-weight and parallel-efficient paragraph representation for MRC based on contextualized representation from long document input, trained using a new Gaussian distribution-based objective that pays close attention to the partially correct instances that are close to the ground-truth. We validate our \systemname{} architecture showing experimental results on two benchmark MRC datasets that require long context modeling; one Wikipedia-based (Natural Questions (NQ)) and the other on TechNotes (TechQA).  \systemname{} can achieve comparable performance on NQ with a state-of-the-art (SOTA) complex document modeling approach while being 16 times faster, demonstrating the efficiency of our proposed model. We also demonstrate that our model can also be effectively adapted to a completely different domain -- TechQA -- with large improvement over a model fine-tuned on a previously published large PLM.
\end{abstract}
\blfootnote{$^\dag$ Equal contributions.}

\section{Introduction}

Machine Reading Comprehension (MRC) has seen great advances in recent years with the rise of pre-trained language models (PLM) \cite{devlin-etal-2019-bert,liu-roberta-2019,lan2019albert} and public leaderboards \cite{Rajpurkar_2016,rajpurkar2018know,yang2018hotpotqa,joshi-etal-2017-triviaqa,welbl-etal-2018-wikihop,kwiatkowski-etal-2019-natural}.  While some challenges \cite{Rajpurkar_2016,rajpurkar2018know} focus on reading comprehension with shorter contexts, many others \cite{welbl-etal-2018-wikihop,joshi-etal-2017-triviaqa,kwiatkowski-etal-2019-natural,DBLP:journals/corr/abs-2101-11272} focus on longer contexts that cannot fit into a typical 512 sub-token transformer window.  Motivated by this, we focus on reading comprehension with long contexts.

One newer approach to this task \cite{zheng-etal-2020-document-acl} focuses on modeling document hierarchy to represent multi-grained information for answer extraction.  Although this approach creates a strong representation of the text, it suffers from a significant drawback. The graph-based methods \cite{velivckovic2018graph} are inefficient on parallel hardware, such as GPUs, resulting in slow inference speed \citep{zhou-2018-gnn,zheng-etal-2020-document-acl}.
Motivated by this, in this paper, we propose a reading comprehension model that addresses the above issue and uses a more light-weight, parallel-efficient (i.e. efficient on parallel hardware) paragraph representation based on long contextual representations for providing paragraph answers to questions. Instead of modeling document hierarchy from tokens to document pieces, we first introduce a base model that builds on top of a large ``long-context" PLM (we use Longformer, \citealp{beltagy-longformer-2020}) to model longer contexts with lightweight representations of each paragraph.  We note that while our approach could work with any PLM, we expect it to perform better with models that can support long contexts and therefore see more paragraph representations at once~\cite{gong-etal-2020-recurrent}.  To provide our model a notion of paragraph position relative to a text we also introduce position-aware paragraph representations (PAPR) utilizing special markup tokens and provide them as input for efficient paragraph classification.  This approach allows us to encode paragraph-level position in the text and teach the model to impute information on each paragraph into the hidden outputs for these tokens that we can exploit to determine in which paragraph the answer resides.  We then predict the answer span from this identified paragraph.

While previous MRC methods \cite{chen-etal-2017-reading,devlin-etal-2019-bert} use ground-truth start and end span positions exclusively as training objectives when extracting answer spans from the context and consider all other positions as incorrect instances equally. However, spans that overlap with the ground-truth should be considered as partially correct. Motivated by \citet{Li-gpo-2020-iclr} which proposes a new optimization criteria based on constructing prior distribution over synonyms for machine translation, we further propose to improve the above base model by considering the start and end positions of ground-truth answer spans as Gaussian-like distributions, instead of single points, and optimize our model using statistical distance.

We call this final model, \systemname{} (\textbf{VA}riable \textbf{U}nified \textbf{L}ong \textbf{T}ext representation) as it can handle a variable number and lengths of paragraphs at any position with the same unified model structure to handle long texts.

\begin{figure}[ht]
     \fbox{\includegraphics[width=\columnwidth]{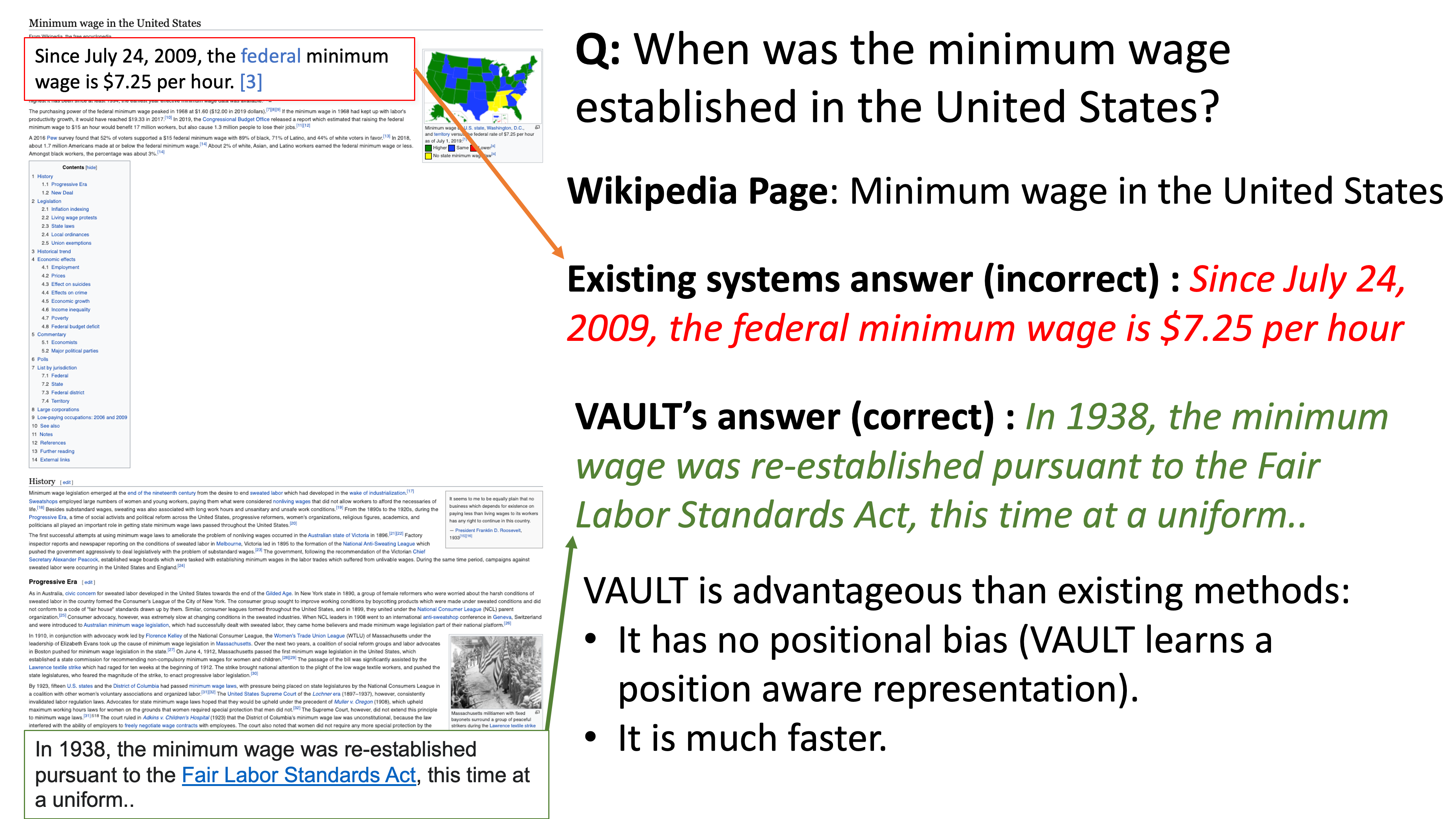}}
     \caption{Example from the NQ dataset with answers from \systemname{} and \cite{zheng-etal-2020-document-acl}.}
     \label{fig:examples}
\end{figure}

To evaluate the performance of \systemname, we select the new Natural Questions (NQ, \citealp{kwiatkowski-etal-2019-natural}) and TechQA \cite{Castelli-techqa-2020-acl} datasets.  NQ attempts to make Machine Reading Comprehension (MRC) more realistic by providing longer Wikipedia documents as contexts and real user search-engine queries as questions, and aims at avoiding \textit{observation bias}: high lexical overlap between the question and the answer context which can happen frequently if the question is created after the user sees the paragraph \cite{Rajpurkar_2016,rajpurkar2018know,yang2018hotpotqa,NQ_COLING,karpukhin-etal-2020-dense,lee2019latent,murdock_et_al_2018_engineered_ai}.
The task introduces the extraction of long answers (henceforth LA; typically paragraphs) besides also requiring short answers (henceforth SA) similar to SQuAD \cite{Rajpurkar_2016}.
In Figure \ref{fig:examples} we examine an example from NQ along with the answers of \systemname{} and \cite{zheng-etal-2020-document-acl}.  We see that while \systemname{} can extract answers from the very bottom of a page -- if relevant -- the existing system suffers from positional bias.  It often predicts answers from the first paragraph of Wikipedia (a region which often contains the most relevant information).  
We evaluate our model for domain adaptation on TechQA, a recently introduced challenging dataset for QA on technical support articles where answers are typically 3-5 times longer 
than standard MRC datasets
\cite{Rajpurkar_2016,rajpurkar2018know}.

Empirically we first show that \systemname{} achieves comparable performance on NQ with \cite{zheng-etal-2020-document-acl}'s document modeling architecture based on graph neural networks while being 16 times faster, demonstrating the efficiency of our proposed model.  Secondly, we show the generalization of our model architecture for domain adaptation on
TechQA.  Our experiments show that our model pre-trained on NQ can be effectively adapted to TechQA outperforming a standard fine-tuned model trained on a large PLM such as RoBERTa.
\noindent
To summarize, our contributions include:

\begin{enumerate}\setlength\itemsep{-0.5mm}
    \item We introduce a novel and effective yet simple paragraph representation.
    \item We introduce soft labels to leverage information from local contexts near ground-truth during training which is novel for MRC.
    \item Our model provides similar performance to a SOTA system on NQ while being 16 times faster and also effectively adapts to a new domain: TechQA.
\end{enumerate}
\section{Related Work}
Machine reading comprehension has been widely modeled as cloze-type span extraction \cite{chen-etal-2017-reading,cui_2017,devlin-etal-2019-bert}. In NQ, we need to identify answers in two levels, long and short answers. \cite{albert-synth-data} adapt a span extraction model for short answer extraction. \cite{zheng-etal-2020-document-acl,liu-etal-2020-rikinet} construct complex networks for paragraph-level representation to enhance long answer classification along with span extraction for short answers. In this work, we propose a more light-weight and parallel-efficient way for constructing paragraph-level representation and classification by using longer context and modeling the negative instance through Gaussian prior optimization.

Using the hierarchical nature of a long document for question answering has been previously studied by \cite{choi-etal-2017-coarse-to-fine-long-documents}, where they use a hierarchical approach to select candidate sentences and extract answers in those candidates. However, due to the limit of input length for large PLMs, existing methods \citep{Alberti-bert-baseline-2019,zheng-etal-2020-document-acl,NQ_COLING} slice long documents into document pieces and perform prediction for each piece separately. In our work, we show that by modeling longer input with position-aware paragraph representation coupled with Gaussian prior optimization (which is novel for MRC), we can achieve comparable performance using much simpler architecture compared to previous models, which coincide with recent new PLM for long inputs on question answering  \cite{Ainslie-ETC-2020}\footnote{The code and model weights of ETC has not been released at the time of writing of the paper for us to have an accurate comparison.}. 
\section{Model Architecture}
\label{sec:models}

\begin{figure*}[ht]
    \centering
     \fbox{\includegraphics[width=\textwidth]{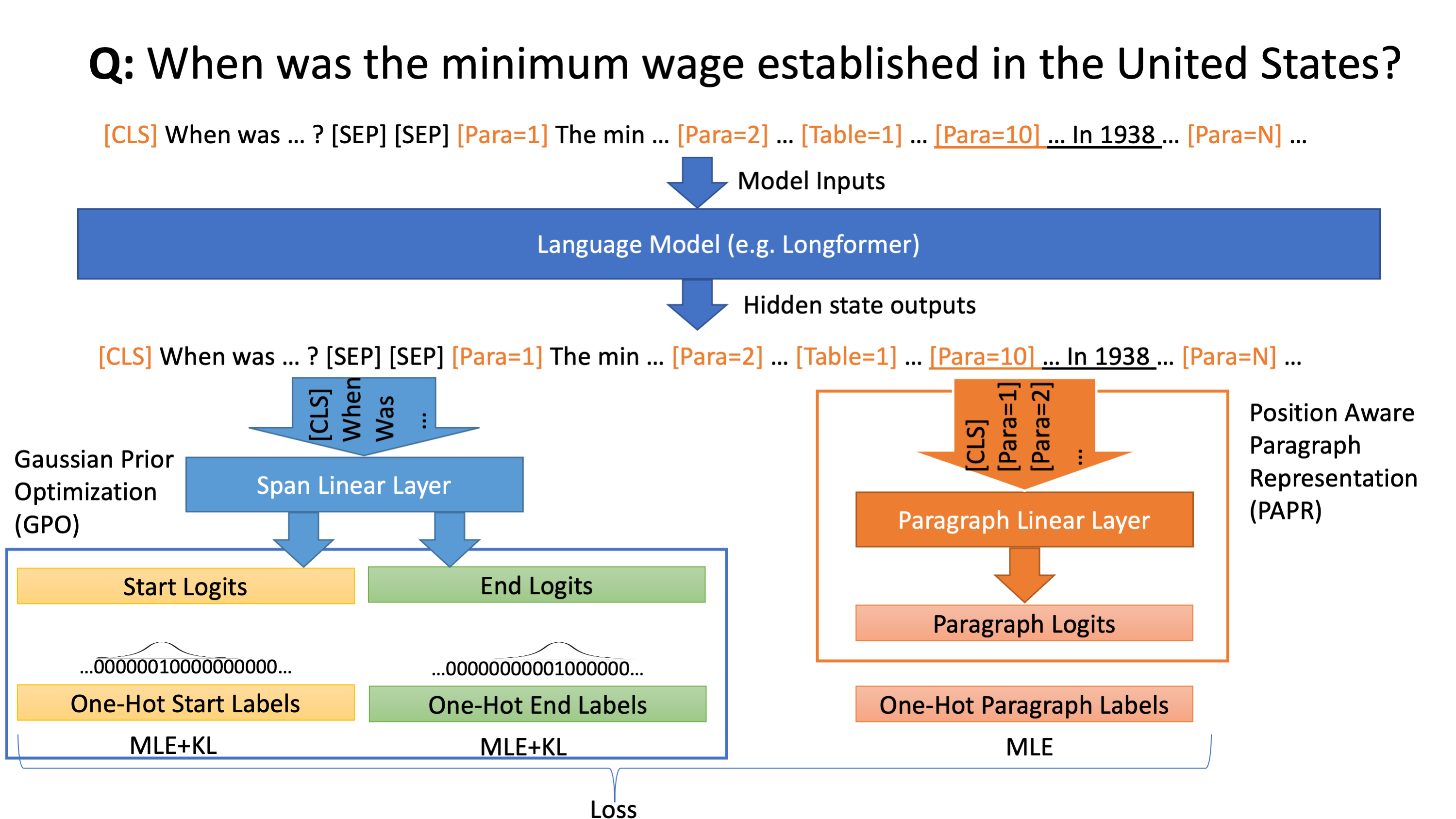}}
     \caption{Overview of \systemname{} answering the example from Figure~\ref{fig:examples}.  The 10th paragraph containing the correct answer is underlined.  The span linear layer receives hidden state outputs from all 4096 tokens in the window to create the start and end logits.  The paragraph linear layer receives the orange-highlighted [CLS] and markup tokens (e.g. [Para=10]) to predict in which paragraph the answer resides.  These logits are then used together to first select the best paragraph (LA) and finally select the best answer within said paragraph (SA).}
     \label{fig:overview}
\end{figure*}

In this section, we introduce \systemname{}, our proposed model that uses a simple yet effective paragraph representation based on a longer context. \systemname{} starts from a base classifier that utilizes position-aware paragraph representations trained on top of a large PLM: Longformer \cite{beltagy-longformer-2020}. Next, we further introduce our Gaussian Prior-based training objective that considers partial credits for positions near the ground-truth, instead of only focusing on one ground-truth position. We show an overview of \systemname{} on the example from Figure \ref{fig:examples} in Figure \ref{fig:overview}.  

\subsection{A Base ``Paragraph'' Predictor Model}
SOTA methods for paragraph prediction \cite{zheng-etal-2020-document-acl, liu-etal-2020-rikinet} represent paragraphs through expensive graph modeling, making it inefficient for ``large-scale'' production MRC systems.  On the other hand, simply selecting the first paragraph performs poorly \cite{kwiatkowski-etal-2019-natural}. We hypothesize that by modeling a much longer context even simple paragraph representation can be effective for paragraph (i.e., long answers) classification. 
For this purpose, we employ a large-window PLM: Longformer \cite{beltagy-longformer-2020}, which has shown effectiveness in modeling long contexts for QA 
\cite{yang2018hotpotqa,welbl-etal-2018-wikihop,joshi-etal-2017-triviaqa}. Compared to conventional Transformer-based PLMs e.g. RoBERTa \cite{liu-roberta-2019} that can only take up to 512 sub-word tokens, Longformer provides a much larger maximum input length of 4,096. 

\noindent \textbf{Position-aware Paragraph Representation (PAPR)}: 
To address the fact that many popular \textit{unstructured} texts such as Wikipedia pages
have relatively standard ways of displaying 
certain relevant information (e.g. birthdays are usually in the first paragraph vs. spouse names are in the ``Personal Life'' paragraph), we provide the base model with a representation of which part of the text it is reading by marking the paragraphs with special atomic markup tokens (\texttt{[paragraph=i]}) at the beginning of each paragraph, indicating the position of the paragraph within the text\footnote{Similar tags are added for lists and tables.}.
With this input representation, we then directly perform long answer classification using the special paragraph token output embedding. Formally, for every paragraph $l_i \in P$, where $P$ are all paragraphs in a text and the representation for the corresponding markup token $\boldsymbol{h}^p_i$, the logit of a paragraph answer $a$ it computes is as $a^p_i = \boldsymbol{W}\boldsymbol{h}^p_i + \boldsymbol{b}$.

We obtain additional document-piece representation from the standard \texttt{[CLS]} \cite{devlin-etal-2019-bert} token to model document pieces that do not contain paragraph answers. The probability of choosing the paragraph given context $\boldsymbol{c}$, is computed as the softmax over paragraph candidate (with an answer span) logits and not containing answer logit: 
\vspace{-1em}
\[
p_l(l_i\mid \boldsymbol{c}) = \text{softmax}(a^p_i).
\]
We pad the paragraph representations to ensure a rectangular tensor in a batch.
Our final prediction strategy is similar to \citet{zheng-etal-2020-document-acl} as we first choose the paragraph candidate with the highest logit among all candidates. We then extract span answers within the selected paragraph answer candidate using a standard pointer network.

\subsection{Gaussian Prior Optimization (GPO)}
Conventional span extraction models \cite{chen-etal-2017-reading, glass-etal-2020-span, liu-etal-2020-rikinet} optimize the probability of predicted start and end positions of the answer spans with ground-truth spans via maximum likelihood estimation (MLE,\citeauthor{wilks-1938-mle}, \citeyear{wilks-1938-mle}). MLE methods promote the probability for the ground-truth positions while suppressing the probability for all other positions. However we hypothesize that, for all those negative instances, the positions that are near the ground-truth should be given higher credit than farther distant positions, since the extracted answers will be partially overlapping with the ground-truth.

To tackle this problem, we follow the intuition from \citet{Li-gpo-2020-iclr} which proposes to promote the probability of generating synonyms using a Gaussian-like distribution for machine translation. We construct the distribution where it has the highest probability at ground-truth positions, and drop the probability exponentially as computed by the distance to the corresponding ground-truth positions. Specifically, for a groundtruth start or end position at $y_s$, where $s\in\{\text{start}, \text{end}\}$, we use a Gaussian distribution $\mathcal{N}(y_s, \sigma)$, where the mean is the position $y_s$ and variance $\sigma$ is a hyperparameter. We consider the probability density $\varphi(y\mid y_s, \sigma)$ of the Gaussian distribution at each position $y$ as the logit for the corresponding position. We then use the softmax function with temperature $T$ to re-scale the logits to get the Gaussian-like distribution $q(y\mid \hat{y}_s)$ for ground-truth distribution at position $y_s$, 
\[
q(y\mid y_s) = \text{softmax}(\varphi(y\mid y_s, \sigma)/T).
\]
We augment our MLE objective with an additional KL divergence \citep{kullback-1951-kld} term between constructed distribution $q(y\mid y_s)$ and model prediction $p_s(y\mid \boldsymbol{c}), s\in\{\text{start}, \text{end}\}$, so that we can guide our model to follow the Gaussian-like distribution for partial credit.
\begin{align*}
L_D \quad=\quad& KL\left(q(y\mid y_s)\parallel p_s(y\mid \boldsymbol{c}) \right) \\
\quad=\quad& \sum_y q(y\mid y_s)\log p_s(y\mid \boldsymbol{c})\\
-& \sum_y q(y\mid y_s)\log q(y\mid y_s).
\end{align*}
We refer to this final model as \systemname{}.
\section{Experiments}

\textbf{Datasets}: We experiment with two challenging ``natural'' MRC datasets: NQ \cite{kwiatkowski-etal-2019-natural} and TechQA \cite{Castelli-techqa-2020-acl}. We provide a brief summary of the datasets and direct interested readers to the corresponding papers.
NQ consists of crowdsourced-annotated \textit{full} Wikipedia pages which appear in Google search logs with two tasks: the start and end offsets for the short answer (SA) and long answer (LA, eg. paragraph) -- if they exist.
TechQA is developed from real user questions in the customer support domain where each question is accompanied  by  50  documents  -- at  most  one of which has an answer -- with answers significantly longer 
(\textasciitilde 3-5x) than standard MRC datasets like SQuAD. We report official F1 scores for each dataset. 
\noindent \textbf{Results on NQ}: We train \systemname{} on NQ -- predicting the paragraph and span answers as NQ's LA and SA respectively -- and compare against \dm{}: a RoBERTa \cite{liu-roberta-2019} variant of the SOTA document model (DM) \cite{zheng-etal-2020-document-acl} 
using the base variants
for a more systematic comparison.  Although it may seem fair to include a Longformer DM baseline in our table, doing so would be infeasible (and unwise) due to production resource constraints.
We further show the impact of \systemname{} by providing ablation experiments where its components (GPO and PAPR) are removed.  The base LM (Longformer in our experiments) without GPO and PAPR, is implemented in the style of \cite{Alberti-bert-baseline-2019,NQ_COLING} where we first predict the SA and then select the enclosing LA.  We aim to show that our proposed method provides comparable results to \dm{} while being considerably faster while decoding and displaying improved performance over experiments just using the language model.  To do this we consider development set SA and LA F1 (the F1 metrics with respect to the span and paragraph answers respectively)
as well as decoding time $t_\text{decode}$ (on a V100) as metrics.

Table \ref{tab:NQ} shows the results on the NQ dev set.  We see \systemname{} and \dm{}  provide comparable F1 performance (precision and recall are shown in the Appendix). However, when it comes to decoding time, we can find \systemname{} decodes over 16 times faster than \dm{}. We additionally see in the ablation experiments that our enhancements increase both F1 metrics by multiple points, at the expense of some decoding time.  In particular we note that the F1 performance of Longformer is not competitive with \systemname{}.  We conclude that \systemname{} provides the best balance of F1 and decoding time as it is effectively tied on F1 (with \dm{}) and is only around 20 minutes slower to decode than the quickest model.


\begin{table}[h]
\centering
\small
    \begin{tabular}{l|cc|c}
\hline
   \textbf{Model} & \textbf{SA F1} & \textbf{LA F1} &  $\boldsymbol{t_\text{decode}}$  \\
   \hline
  \dm & \textbf{52.2} & 70.1 & 11h   \\
\hline
   \systemname  & 51.6  & \textbf{70.4} & 40m \\
\hline
   - GPO & 49.1 & 67.6  & 41m \\
   - PAPR (Longformer) & 49.5 & 65.6 & \textbf{22m}   \\
   \hline
\end{tabular}
\caption{Comparison of \systemname{} vs. \dm \ on NQ. We achieve comparable performance while being 16 times faster.}
  \label{tab:NQ}
\end{table}


\noindent \textbf{Domain Adaptation: Results on TechQA}: Since \systemname{} has shown to be effective on NQ, we evaluate it on a new domain, TechQA.  We compare it against a RoBERTa base model trained with the same hyper-parameters as \cite{Castelli-techqa-2020-acl} -- except we use 11 epochs instead of 20.  We chose base instead of large (as is used for the TechQA baseline) to give a fair comparison since we are using a base PLM for our experiments with \systemname.  Similarly, we use RoBERTa rather than BERT as it is closer to Longformer. 
Having already established the run-time effectiveness of \systemname{} on NQ, we focus on F1 metrics here, including ``has answer'' (HA) F1.  We consider HA F1 our primary metric as we are exploring paragraph answer extraction in this work and (as previously mentioned) answers in TechQA are much longer than other datasets. 
We believe that the improvements in HA F1, at least partially, come from GPO.

\begin{table}[h]
\centering
\small
\begin{tabular}{l|cc}
\hline
 
   \textbf{Model} & \textbf{F1} & \textbf{HA F1} \\
   \hline
  RoBERTa & 48.6 & 7.6    \\
\hline
   \systemname & \textbf{49.3}  & \textbf{16.1}    \\
   \hline
\end{tabular}
\caption{Results on TechQA dev set. \systemname{} clearly outperforms RoBERTa on both F1 and Has Answer F1.}
  \label{tab:TechQA}
\end{table}

Results on TechQA are reported in Table \ref{tab:TechQA}.  We see that our \systemname{} model provides an improvement of 0.7 F1 and 8.5 HA F1 (denotes Has Answer); thus showing the effectiveness of our approach.  In particular, we see that this approach of imputing a paragraph structure to classify provides a large boost to performance when a non-null answer exists (HA F1).
\section{Conclusions}
In this work we introduce and examine a powerful yet simple model for reading comprehension on long texts which we call \systemname, based on the hypothesis that with a large sequence length long answers can be classified effectively without computationally heavy graph-based models.  We validate our approach by showing it yields F1 scores competitive with heavier methods at a fraction of the decoding cost on two very different domain benchmark datasets that require reading long texts.




\bibliography{anthology,vault}
\bibliographystyle{acl_natbib}

\appendix

\section{Additional Experimental Results}

\begin{table*}[ht]
\centering
\small
    \begin{tabular}{l|ccc|ccc}
\hline
   \textbf{Model} & \textbf{SA F1} & \textbf{SA P} & \textbf{SA R} & \textbf{LA F1} & \textbf{LA P} & \textbf{LA R}   \\
   \hline
  \dm{} & \textbf{52.2} & 57.2 &  \textbf{48.0} & 70.1 & 69.4 & 70.9  \\
\hline
   \systemname{} & 51.6 & \textbf{61.5} & 44.4  & \textbf{70.4} & \textbf{69.5} & \textbf{71.4} \\
\hline
   - GPO & 49.1 & 57.6 & 42.7 & 67.6  & 67.0 & 68.1 \\
   - PAPR (Longformer) & 49.5 & 56.4 & 44.2 & 65.6 & 62.4 & 69.3   \\
   \hline
\end{tabular}
\caption{Comparison of \systemname{} vs. \dm{} on NQ with precision (P) and recall (R) statistics.}
  \label{tab:NQ-extended}
\end{table*}

For interested readers we further show precision and recall numbers for the NQ experiments in Table~\ref{tab:NQ-extended}.

\section{Implementation Details}
\subsection{NQ}
All models for this work are implemented in \cite{Wolf2019HuggingFacesTS}.
We use the following hyperparameters for \systemname{} when finetuning on NQ: sequence length 4096, doc stride 2048 \cite{Ainslie-ETC-2020}, negative instance subsampling rates (has answer/no answer) 0.02/0.08, learning rate 5e-5, and 4 epochs of training.

\subsection{TechQA}
While TechQA does provide full HTML for its Technotes, the answers are annotated with respect to the cleaned plaintext.  Therefore to determine paragraph breaks for \systemname{} we split on the \verb|"\n\n"| token \verb|"ĊĊ"| in the vocabulary.  By imputing paragraph answers in this way, we are then able to predict the paragraph answer and then a contained span answer. 

\section{Example Analysis}

\begin{figure}[ht]
    \centering
    \small
\framebox{%
  \begin{minipage}{\columnwidth}
    \underline{\textbf{Example A1 (NQ)}} \\
\textbf{Question:} why did government sponsored surveys and land acts encourage migration to the west \\
\textbf{Wikipedia Page:} Homestead Acts \\
\textbf{Text:} ... \\
\textcolor{hi-green}{An extension of the Homestead Principle in law, the Homestead Acts were an expression of the " Free Soil " policy of Northerners who wanted individual farmers to own and operate their own farms, as opposed to Southern slave-owners who wanted to buy up large tracts of land and use slave labor, thereby shutting out free white men.} \\ \\
\textcolor{hi-red}{The first of the acts, the Homestead Act of 1862 , opened up millions of acres. Any adult who had never taken up arms against the U.S. government could apply. Women and immigrants who had applied for citizenship were eligible. The 1866 Act explicitly included black Americans and encouraged them to participate, but rampant discrimination slowed black gains. Historian Michael Lanza argues that while the 1866 law pack was not as beneficial as it might have been, it was part of the reason that by 1900 one fourth of all Southern black farmers owned their own farms. [1]} \\
... \\
\\
\underline{\textbf{Example A2 (TechQA)}} \\
\textbf{Question:} Are there any probes that can connecto the Nokia NSP EPC v17.9 and Nokia NSP RAN v17.3 using JMS/HTTP?\\
\textbf{Text:} release notice; downloads; nco-p-nokia-nfmp; Probe for Nokia Network Functions Manager for Packet NEWS \\ \\ 
ABSTRACT \\ This new probe will be ready for downloading on July 20, 2017. \\ \\ 
CONTENT \\ \\ \\
\textcolor{hi-green}{This probe is written to support Nokia Network Functions Manager for Packet release 17.3}. \\ \\
\textcolor{hi-red}{You can download the package you require from the IBM Passport Advantage website}: \\ \\ www-01.ibm.com...\\
  \end{minipage}}    
  \caption{Additional Examples of questions in the NQ and TechQA datasets.  \systemname{}'s correct answer is shown in green,  incorrect baseline in red. (A1) The correct answer is a paragraph LA; only \systemname{} identifies the correct LA directly even though the gold SA is null.  (A2) \systemname{} identifies the correct "paragraph" answer.}
    \label{fig:examples2}
\end{figure}
We examine additional examples in Figure \ref{fig:examples2} to provide insight on the improvements of \systemname{}.  We compare the correct answers produced by \systemname{} with the incorrect answers produced by the ablated model from the last row of Table~\ref{tab:NQ-extended} (NQ) and Roberta baseline from the first row of Table~\ref{tab:TechQA} (TechQA).

In the first example the gold SA is null, however there is a gold LA.  This indicates that there is no short span which answers the question: the correct answer here is an entire paragraph.  This does not confuse \systemname{} which is able to identify the correct answer directly.  However the ablated model which attempts to predict SA first struggles here -- predicting the incorrect LA -- as there is no gold SA.

In the second example we see that in this Technote both the correct and incorrect answers are single sentence paragraphs surrounded by paragraph breaks.  Our \systemname{} is able to identify the correct paragraph using our imputed structure and select the correct answer -- whereas the Roberta baseline selects a nearby but incorrect answer. 

\end{document}